\documentclass[letterpaper, 10 pt, conference]{ieeeconf}  
\usepackage{graphicx}

\IEEEoverridecommandlockouts                              

\usepackage{amsmath} 
\usepackage{amsfonts}  
\usepackage{array} 
\usepackage{multirow} 
\usepackage{caption}
\usepackage{booktabs}
\usepackage{commath}
\usepackage{mathtools}
\usepackage{hyperref}
\usepackage{gensymb}
\usepackage{xcolor}
\usepackage{lipsum}
\usepackage{graphicx}
\usepackage{textcomp}
\usepackage{subcaption}

\usepackage[allow-number-unit-breaks]{siunitx}

\usepackage{placeins}

\title{\LARGE \bf
Side Scan Sonar-based SLAM for Autonomous Algae Farm Monitoring
}

\author{Julian Valdez$^{1}$, Ignacio Torroba$^{1,2}$, John Folkesson$^{1}$ and Ivan Stenius$^{2}$
\thanks{The authors are with $^{1}$ the \href{https://www.kth.se/is/rpl/division-of-robotics-perception-and-learning-1.779439}{Division of Robotics, Perception and Learning} and $^2$ the Division of Naval Architecture at KTH Royal Institute of Technology, SE-100 44 Stockholm, Sweden.
{\tt\small \{jvaldez, torroba, johnf, stenius\}@kth.se}}}%

\begin{document}

\maketitle
\thispagestyle{empty}
\pagestyle{empty}

\begin{abstract}
The transition of seaweed farming to an alternative food source on an industrial scale relies on automating its processes through smart farming, equivalent to land agriculture. Key to this process are autonomous underwater vehicles (AUVs) via their capacity to automate crop and structural inspections. However, the current bottleneck for their deployment is ensuring safe navigation within farms, which requires an accurate, online estimate of the AUV pose and map of the infrastructure. 
To enable this, we propose an efficient side scan sonar-based (SSS) simultaneous localization and mapping (SLAM) framework that exploits the geometry of kelp farms via modeling structural ropes in the back-end as sequences of individual landmarks from each SSS ping detection, instead of combining detections into elongated representations.
Our method outperforms state of the art solutions in hardware in the loop (HIL) experiments on a real AUV survey in a kelp farm. The framework and dataset can be found at \url{https://github.com/julRusVal/sss_farm_slam}.
\end{abstract}

\section{INTRODUCTION}
Algae has a large potential as a future food source as well as a raw material for fossil-free products \cite{krause2022prospects}.  Several types of seaweed are relatively easy to cultivate and there is a considerable amount of suitable area for it along our coastlines.  However, for it to become a competitive resource, its production must be scaled up efficiently by means of so-called smart farming. While agriculture on land has seen an enormous increase in automation recently through the use of drones \cite{sa2018weedmap} and ground robots \cite{fountas2020agricultural}, marine farming poses several distinctive challenges to the deployment of AUVs \cite{Stenius22}.

\begin{figure}[ht]
    \centering
    \includegraphics[width=0.8\linewidth]{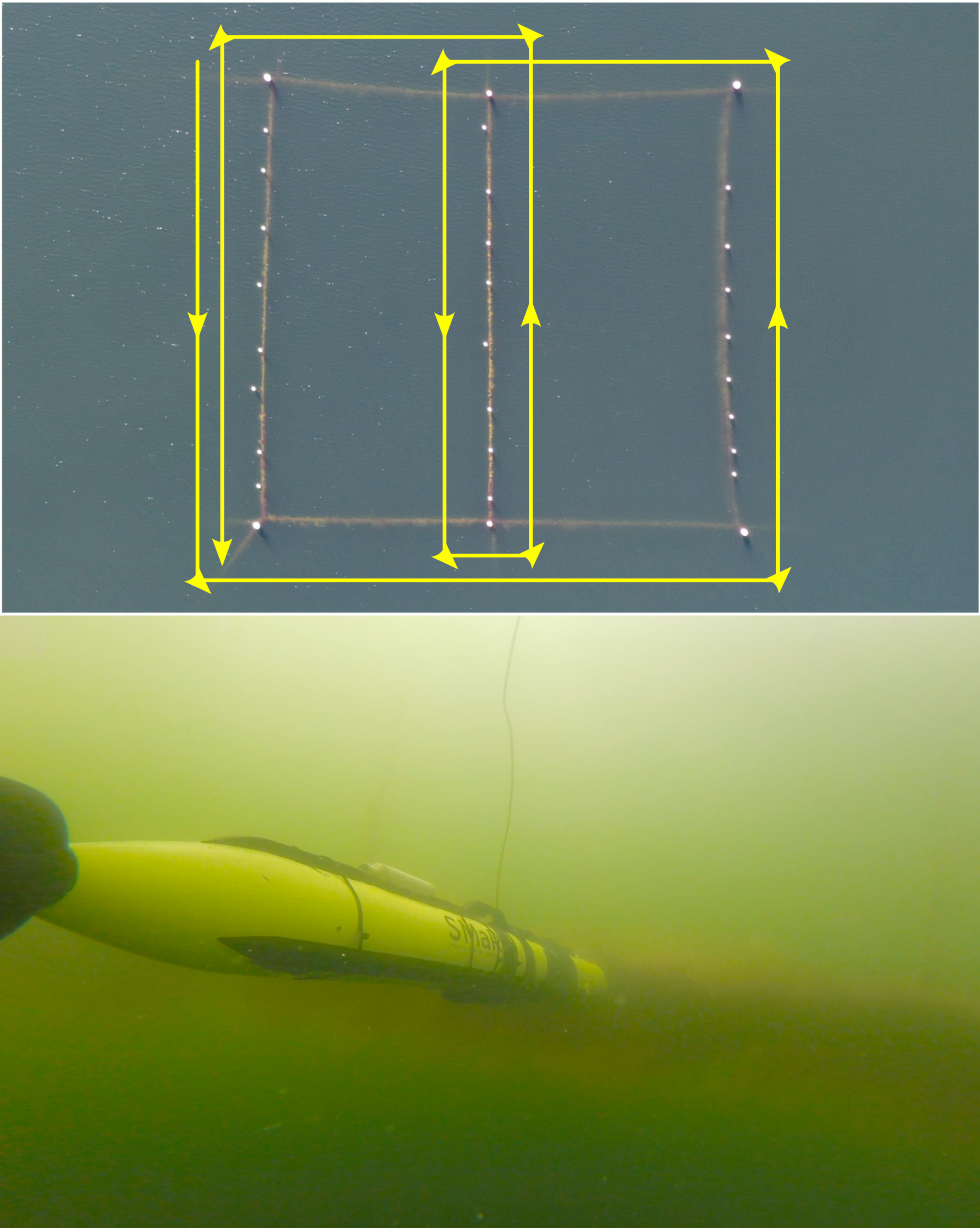}
    \caption{Aerial view of the algae farm (top) with a depiction of the survey plan and the AUV (bottom) used for the survey.}
    \label{fig:farm_real}
\end{figure}

After initially laying out the juvenile algae on ropes between moored buoys (see top of Fig. \ref{fig:farm_real} for an aerial view of an algae farm), marine farming only requires monitoring for several months while the algae grows to harvesting length. 
Farmers need to periodically assess the health and growth of the algae as well as the structural status of the farms, which are prone by design to be disrupted by meteorological events or strong currents. Currently, monitoring operations require a vessel, divers, and crew.  AUV's have been proposed in the literature as a tool to reduce the associated risks and costs of such tasks \cite{utne15}.     
However, the cost-effective deployment of AUVs in algae farms imposes constraints both in the size of the vehicles that can navigate among the seaweed lines and their value. Smaller, more affordable AUV models provide the higher maneuverability required but at the expense of experiencing a larger degradation in their pose estimate over time. Such navigational drift is currently the main bottleneck for the safe deployment of AUVs in algae farms \cite{Stenius22}.  

Solutions involving installing underwater acoustic beacons, such as ultra short baseline (USBL) \cite{Reis2016}, result in increasing costs and complexity. Alternatively, SLAM techniques have proved successful at bounding the AUVs localization drift by means of exteroceptive measurements \cite{fallon2011efficient}.
Thus, in this paper we propose a graph-SLAM method for AUVs equipped with a SSS that utilizes the sonar measurements to both construct a structural model of the farm (top of Fig. \ref{fig:farm_real}) and provide an accurate pose estimate of the vehicle online (bottom of Fig. \ref{fig:farm_real}) for safe navigation.

Our approach exploits the geometry of the farm lines by dealing with the data association problem over rope detections implicitly in the back-end, instead of through feature-matching methods in the front-end. Instead of trying to associate corresponding single rope segments from SSS beams during a loop closure or to a line-shaped prior of the rope, which are unconstrained problems, our approach models each rope segment detection as independent landmarks.
By associating a loose rope prior with high uncertainty along the ropes' directions to each individual new landmark, "sliding" edges are added to the graph. These edges constrain the vehicle's estimate laterally with respect to the ropes while allowing dead reckoning (DR) constraints to anchor the optimization along the direction of the lines. Our experiments show that the resulting sparsely-connected graph of the farm can be optimized incrementally in real time through smoothing techniques such as iSAM2 \cite{isam2}. 

We compare our method against the state of the art techniques presented in \cite{fallon2011efficient} and \cite{zhang2023fully}, in which measurements arising from the same landmarks are associated in the front-end, to show how our treatment of the measurements on the back-end results in a very simple front-end and very sparsely-connected graphs. We evaluate our approach in a sea trial in terms of mapping quality and AUV trajectory estimates.

\section{RELATED WORK}
AUVs have been presented in recent literature as tools to monitor aquaculture farms \cite{bell2020utility}. However, the challenge remains in choosing a robust sensor suite that can be mounted on small, agile vehicles while providing all the necessary information of the environment regardless of the water conditions \cite{Stenius22}.  While the use of camera feedback is advocated for seaweed monitoring in underwater farms in \cite{gerlo2023seaweed}, poor visibility can quickly render visual-inertial navigation systems \cite{joshi2019experimental} unsuitable for underwater SLAM in coastal sea water. To ameliorate the dependence of visual information to water conditions, \cite{Fischell18} proposed combining camera feedback with the use of split-beam sonar data. Other approaches utilize multibeam echosounders for mapping, however these are expensive and require larger AUVs to be mounted. As an alternative, SSS are lighter and inexpensive, although their use in SLAM frameworks remains a challenge due to the difficulty in feature matching on SSS data \cite{fallon2011efficient} and the degeneracy in landmark estimation that stems from the lack of depth information in SSS measurements \cite{zhang2023fully}. 


Regarding feature matching in SSS imagery, while a large amount of deep learning (DL) methods have been recently proposed in the literature \cite{neupane2020review}, the latest solutions for AUV navigation such as \cite{petrich2018side, zhang2023fully} rely on model-based feature extraction and matching for their front-end, such as those compared in \cite{king2013comparison}. In our specific setup, the choice of a model-based approach to feature detection is based on two arguments: i) training a DL architecture requires amounts of data that are not currently available for our specific scenario; ii) the domain-specific knowledge of our scenario allows for the design of target-specific detectors that perform well while requiring low computational resources compared to DL models. This is particularly important for our method to be run in an embedded platform in real time.

Concerning the SSS-SLAM back-end, while EKF-based frameworks exist \cite{siantidis2016side}, \cite{fallon2010cooperative} argues that graph-based solutions to the map and vehicle trajectory estimation are best suited for frameworks based on acoustic measurements, such as those presented in \cite{fallon2011efficient, zhang2023fully}. Both of these methods apply smoothing techniques to compute a maximum a-posteriori (MAP) approximation to the SLAM posterior formulated from combining DR and SSS relative constraints into a non-linear least-squares problem. These are the most closely-related works to ours. However, \cite{fallon2011efficient} requires ground truth associations for feature matching in SSS imagery, as opposed to our method. While in \cite{zhang2023fully} the feature matching is approached programmatically, each detection is associated with the estimated point-landmark from which the measurement is believed to have originated and the back-end optimization is run at the end of each SSS swath. The former does not work well in the presence of elongated landmarks (the ropes in our setup) and the latter would not provide a real-time vehicle estimate accurate enough to avoid collisions in a confined space such as an algae farm. To ameliorate both shortcomings, we present our solution to SSS-based SLAM in algae farms in the next section.



\begin{figure}[ht]
    \centering
    \includegraphics[width=0.9\linewidth]{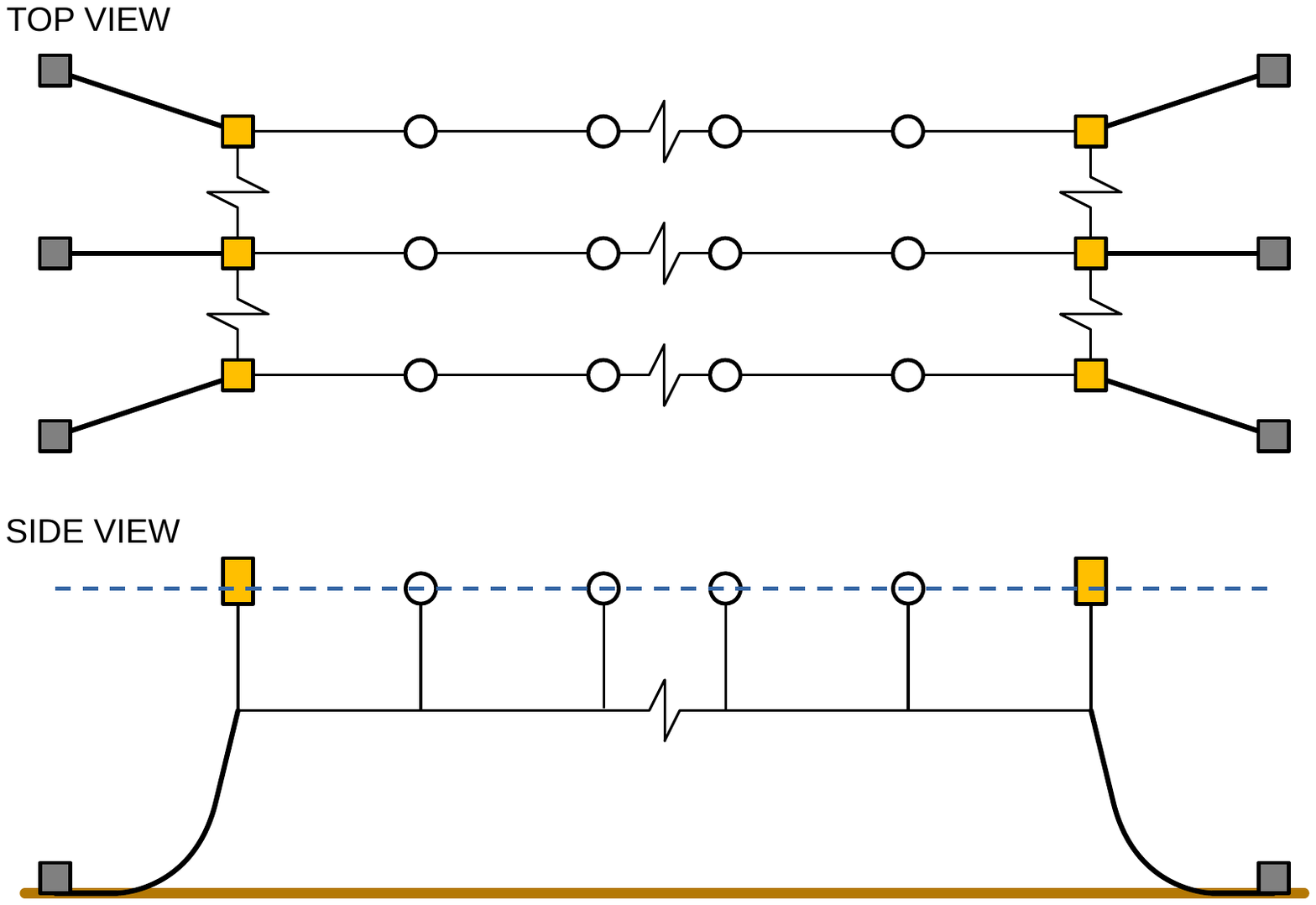}
    \caption{Diagram of an algae farm infrastructure. The algae hangs from the horizontal lines supported by the buoys (mooring buoys in yellow, intermediate buoys in white). The thicker lines and the gray boxes indicate mooring lines and anchors, respectively.}
    \label{fig:farm_structure}
\end{figure}

\section{ALGAE FARM STRUCTURAL MAPPING}
\label{sec:structural_mapping}
A common algae farm structure for Saccharina Latissima (sugar kelp) generally comprises rows of parallel ropes from which the algae grows (see top of Fig. \ref{fig:farm_structure}) and mooring buoys at both ends of the lines that hold them close to the surface and anchored to the seabed (bottom of Fig. \ref{fig:farm_structure}). The rows are held at a relatively consistent depth by intermediate buoys along their length. See \cite{thomas2021comparative} for a more detailed description of the cultivation infrastructure.
As these structures are susceptible to relative displacements from their original configuration due to storms and currents, they must be remapped periodically to assess displacements and determine if actions are required \cite{bell2020utility}.
In this work, similarly to \cite{Stenius22}, we argue that a sufficient model of an algae farm at time $t$ for such tasks can be parameterized via the set of $K$ 2D poses of the mooring buoys $B_t = \{b_{k}\}_{k=1}^{K}$, where $b_k = [x_k, y_k]$, from which the 2D poses of the ropes can be approximated assuming they are straight lines. 


\subsection{AUV motion estimate}
Autonomous inspection of underwater infrastructures with an AUV involves creating a spatial representation of the target structure using exteroceptive measurements collected by the vehicle along its trajectory. This continuous trajectory, defined as the 6-dimensional state of the vehicle $[x_t, y_t, z_t, \phi_t, \psi_t, \theta_t]$ sampled at every time step $t$, can be estimated by means of dead reckoning (DR). 
In the case of AUVs, the depth $z$, roll $\phi_t$, and pitch $\psi_t$ parameters can be directly measured with no significant drift by means of a barometer and an IMU. This allows for the DR estimation to be reduced to three dimensions $m_t = [x_t, y_t, \theta_t]$.
Assuming zero-mean, additive Gaussian noise, the DR of the AUV can be generally expressed as $m_{t} \sim \mathcal{N}(f(m_{t-1}, u_{t}), \Sigma_t)$, where $u_t$ and $f(\cdot)$ are the vehicle's control input and its motion model respectively and $\Sigma_t$ is the covariance matrix.

\subsection{SSS measurement model}
A side scan sonar attached to the AUV provides relative measurements of the buoys $g_j$ which are used to estimate the set of buoy poses $B_t$ and keep the AUV localized. However, the scarcity of the buoys along the farm renders them insufficient as the only mean of autonomous localization. 
For this reason, our SLAM framework utilizes rope detections from SSS imagery as well, given their ubiquity through the survey. Therefore we define the relative, range and bearing, measurements of rope segments as $q_l$. 

\subsubsection{Target detection on SSS imagery}
\begin{figure}[t]
    \centering
    \vspace{0.075in}
    \includegraphics[width=.85\linewidth]{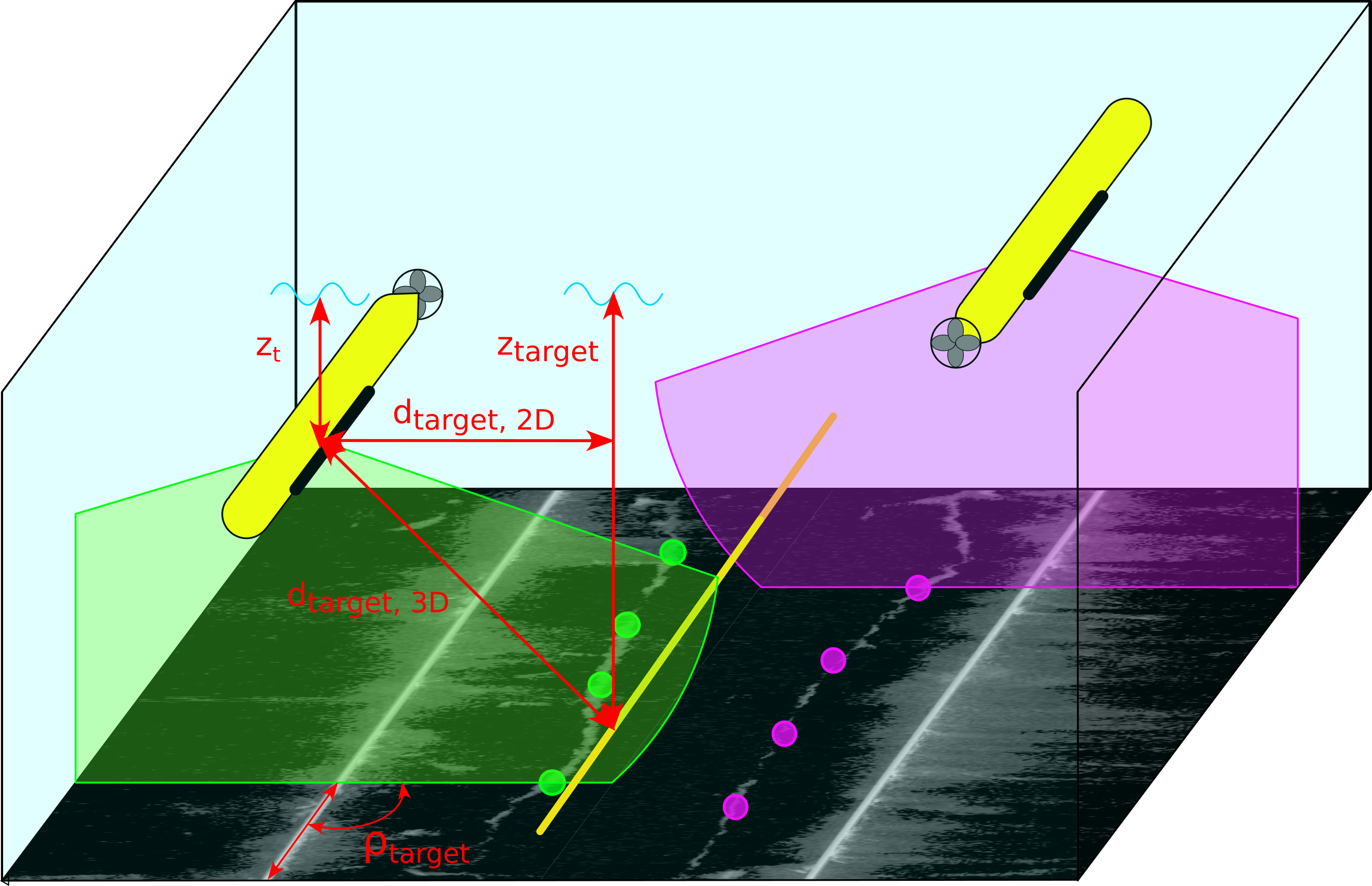}
    \caption{Depiction of two parallel SSS swaths from combining consecutive SSS pings (purple and green) collected along a farm line (yellow). 2D rope detections in the SSS imagery (green and purple circles) are used to extract the rope coordinates through the SSS geometry model, in red. The misalignment in the detections is due to the DR drift.}
    \label{fig:data_association}
\end{figure}

SSS sonars produce 1D pings in a fan-shaped pattern downward from the vehicle (see Fig. \ref{fig:data_association}), which can be combined to form waterfall 2D images as the AUV moves. In this work, however, both the buoys and ropes detections have been extracted online directly from each SSS ping applying the methodology in \cite{Stenius22} and their 2D positions have been mapped to real coordinates through the geometric model of the SSS, introduced below. The final SSS detections can be seen in Fig. \ref{fig:detector_overlay}. 



\subsubsection{Geometric SSS model}
The detections from the SSS pings are used to extract the slant range to the target, $d_{target, 3D}$. Given that the depth of the vehicle can be measured directly $z_t$ and the depth of the target is known a priori, $z_{target}$, the relative 2D range, $d_{target, 2D}$, and bearing, $\rho_{target}$, of a target with respect to the vehicle can be determined as follows, similarly to \cite{fallon2011efficient}, assuming a flat bottom:

\begin{align}
\begin{split}
    d_{target, 2D} &= \sqrt{d_{target, 3D}^2 - {\vert z_t - z_{target} \vert}^2} \\
    \rho_{target} &= \pm \pi/2
    \label{eq:meas_model}
\end{split}
\end{align}

The sign of $\rho_{target}$ is determined by the SSS channel, port or starboard. 
According to Eq. \ref{eq:meas_model}, a relative position constraint is therefore represented by $[d_{target, 2D}, \rho_{target}]$ for both buoys and rope segments, with $z_{buoy} = 0$ and $z_{rope} = 1.5$m in the farm in Fig. \ref{fig:farm_real}. Modeling the uncertainties in the detection processes as zero-mean Gaussian, both measurement models for buoys and ropes can be defined as $g_{j} \sim \mathcal{N}(h(m_j, b_j), \Omega_j)$ and $q_{l} \sim \mathcal{N}(d(m_l, r_l), \Xi_l)$.


\section{Our method}
The vehicle's DR estimate and the SSS measurements of buoys and rope segments introduced above can be combined in a landmark-based graph-SLAM back-end to approximate both the vehicle trajectory and the landmarks positions that best minimize the errors in the survey. However, the data association problem arising from the combination of floating ropes as landmarks and single ping detections from the SSS requires a novel treatment of the feature matching process. In our approach, we couple a simple and fast front-end with a formulation of the rope landmarks in the back-end that allow matching single 2D detections to 3D ropes directly, correcting the AUV trajectory with each new detection. 

\subsection{Data association for rope detections in SSS}
As seen in Fig. \ref{fig:data_association}, the SSS survey swaths collected along the algae lines yield individual per-ping detections of ropes segments. These detections across adjacent swaths can be used to correct for the inertial navigation error of the AUV, depicted in Fig. \ref{fig:data_association} as a misalignment between detections across swaths. However, this is not trivial since individually they are not unambiguous enough, yielding an unconstrained data association problem. The most direct state of the art solution to feature matching with SSS data would be that of \cite{zhang2023fully}, in which individual detections are directly associated to each other. \cite{zhang2023fully} will be tested as a baseline in Sec. \ref{sec:results} and proved to not perform well with elongated features.

A more robust solution must therefore involve associating the 2D SSS detections to the long ropes they originate from, whose initial poses can be modelled by a set of priors. Therefore defining these rope priors adequately is key for a working solution. We have chosen to parameterize them as 2D Gaussian distributions instead of 3D lines. The reason for this is two-fold: i) the online processing of SSS pings yields 2D measurements that cannot be individually matched to a line model (the problem is under-constrained). They could be processed in batches instead (traditionally referred to as submaps in the SLAM literature \cite{aulinas2011robust}) but this yields DR errors within the submaps themselves that cannot be corrected, ii) the algae ropes are not static straight lines, but rather undulate dynamically in the water column due to currents and the weight of the algae, as can be appreciated in Fig. \ref{fig:detector_overlay}. Line priors would fail to capture this. Therefore the rope priors are of the form $P(r_v) \sim \mathcal{N}(r^{prior}_v, \Lambda_v)$, where $r^{prior}$ is computed as the middle 2D coordinate between the two mooring buoys holding a rope and $\Lambda$ models a large uncertainty along the lines and very narrow across them.

\begin{figure}[th]
  \centering

  \begin{subfigure}[]{0.4\textwidth}
    \centering
    \includegraphics[width=\textwidth]{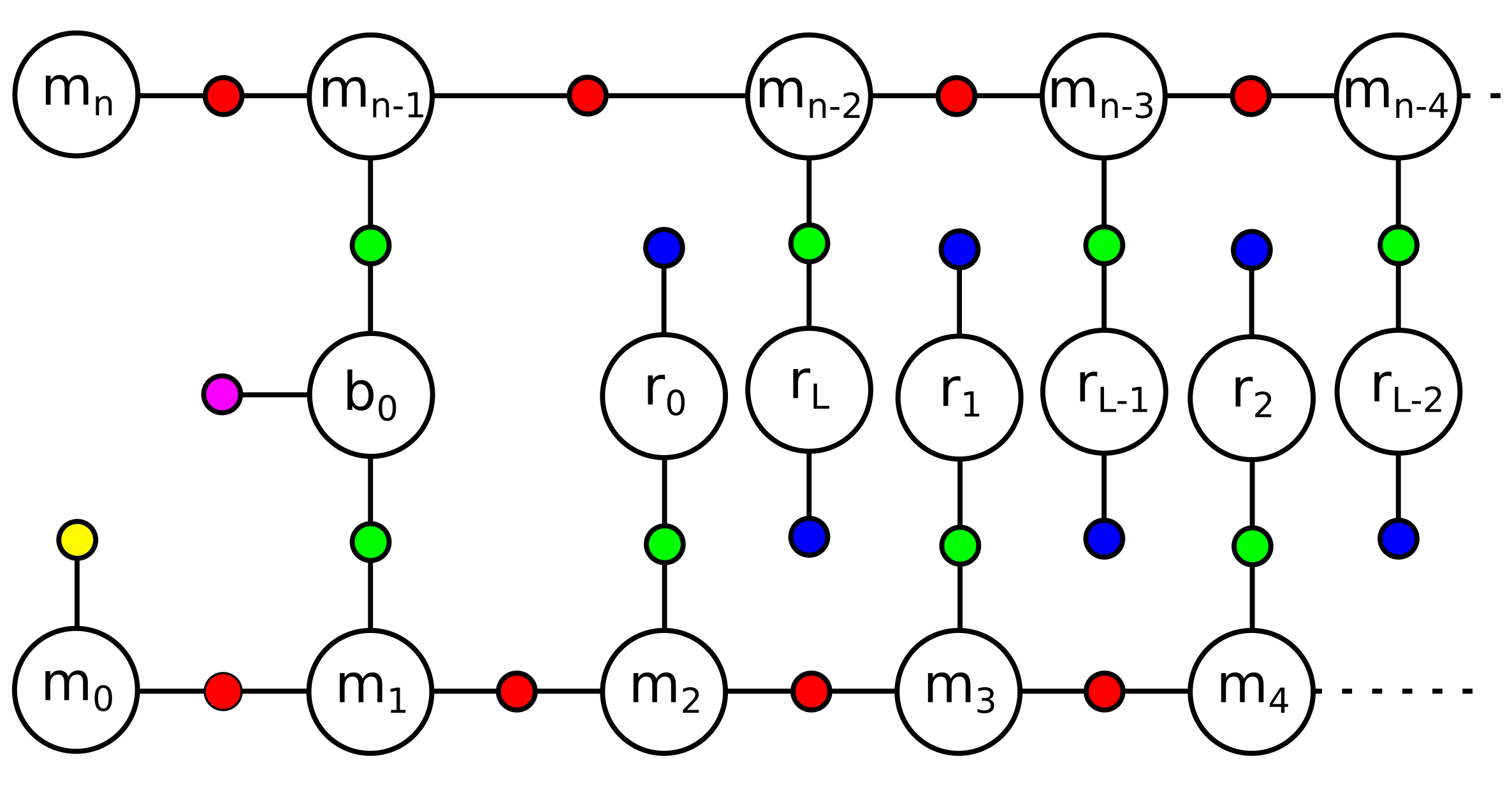}
    \caption{Proposed method}
    \label{fig:factor2}
  \end{subfigure}
  
  \begin{subfigure}[]{0.4\textwidth}
    \centering
    \includegraphics[width=\textwidth]{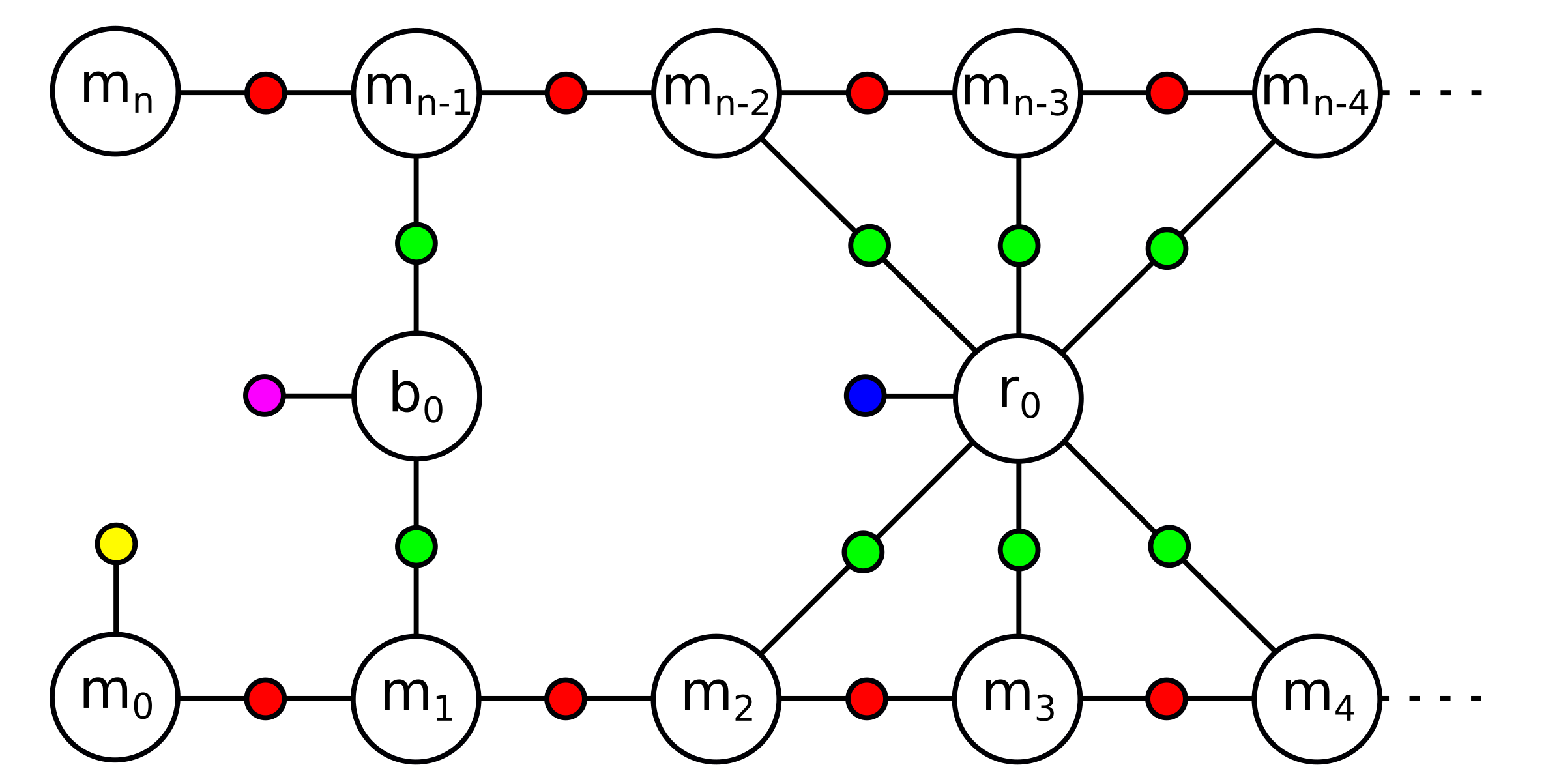}
    \caption{Baseline method \cite{fallon2011efficient}}
    \label{fig:factor 1}
  \end{subfigure}

  \caption{Factor graph formulations of the presented approach (top) and the baseline method from \cite{fallon2011efficient} (bottom). Variable nodes are indicated by white circles: $m_t$ represents the AUV states, $b_j$ represents the buoy positions, and $r_l$ represents the rope positions. Factor nodes are indicated by colored circles: yellow for priors on the initial state of the AUV, magenta for priors on buoys, blue for priors on ropes, red for odometry measurements, and green for detection measurements. }
  \label{fig:factor_graphs}
\end{figure}

With this prior model, the data association problem is now reduced to matching each 2D detection to the 2D prior of the rope from which they originated, which is trivial in our setup given the AUV pose estimate and the prior map of the farm. The factor graph of this approach can be seen in Fig. \ref{fig:factor 1} and its equivalent to that from \cite{fallon2011efficient}, which will be used as another baseline method. However, despite the large along-line uncertainty in the priors, this alone would result in the graph corrections diverging rapidly with the first detections of a new line (this will be shown in Sec. \ref{sec:results}). To prevent this, we propose instead treating all rope detections as arising from different rope landmarks, thus attaching each detection to a new rope prior instance, with the priors being the same for every detection of the same physical line. This approach yields the factor graph in Fig. \ref{fig:factor2}. This method generates a more densely populated but more sparsely connected factor graph that can thus be optimized faster. Furthermore, it also results in a very simple front-end. But the key to its good performance lies in the parameterization of the prior uncertainties $\Lambda$, whose principal components are proportional to the ropes, making them very simple to tune. This results in SSS factors that allow the $m_t$ estimates to "slide" in parallel to the ropes during the graph optimization, driven by the DR factors, while firmly anchoring the lateral distances of the AUV poses wrt the ropes, which are measured accurately with the SSS.

\subsection{Graph optimization}
The joint probability distribution of the vehicle trajectory and the measurements of the farm up to $t = \Delta$ from the Bayes nets associated to the graphs in Fig. \ref{fig:factor_graphs} is given by: 
\begin{multline}
\label{eq:joint_prob}
    P(M, U, G, Q) = P(m_0)  \prod_{t=1}^{\Delta} P(m_t \vert m_{t-1}, u_t) \prod_{k=0}^{K} P(b_k)  \\ 
    \prod_{v \in V}^{W} P(r_v) \prod_{(t,k) \in G}^{J} P(g_{tk} \vert m_t, b_k)  
    \prod_{(t,s) \in Q}^{L} P(q_{ts} \vert m_t, r_s)
\end{multline}

Where $G = \{g_{tk}\}^{J}$ and $Q = \{q_{ts}\}^{L}$ are the sets of buoys and ropes detections respectively up to time $t=\Delta$. The matching of the $J$ buoy detections to the $K = 6$ buoys and the $L$ rope segment detections to the corresponding $S = 3$ ropes has been done through a maximum likelihood procedure. $P(m_0) \sim \mathcal{N}(m_0, \Sigma_0)$, $P(b_k) \sim \mathcal{N}(b^{prior}_k, \Gamma_k)$ and $P(r_v) \sim \mathcal{N}(r^{prior}_v, \Lambda_v)$ are the priors on the first vehicle pose, the buoys and the ropes, respectively.
Additionally, $V$ is the set of rope indexes corresponding to the $W$ rope priors added to the net.
Note that the main difference between our method and that of \cite{fallon2011efficient} in Eq. \ref{eq:joint_prob} is in the number of rope priors $W$ used. While in our method $W = L$, in that of \cite{fallon2011efficient} $W = S$, the number of real ropes in the farm.

Given the $L_{\Delta}$ rope, $J_{\Delta}$ buoy detections and the control inputs $U=\{u_t\}_{t=0}^{\Delta}$, we can estimate the sets of latent variables $M=\{m_t\}_{t=0}^{\Delta}$, which is the vehicle trajectory up to $t=\Delta$ the buoys $B=\{b_k\}_{k=0}^K$ and the ropes positions $R=\{r_v\}_{v=0}^W$ through a maximum a posteriori (MAP) estimate of Eq. \ref{eq:joint_prob}, which results in the following non-linear least-squares problem:
\begin{multline}
\label{eq:map}
    M^*, B^*, R^* = \underset{M,B,R}{\text{argmin}}\; \sum_{t=1}^{\Delta} {\parallel f(m_{t-1}, u_t) - m_t \parallel}_{\Sigma_t}^{2} \\
    + \sum_{k=0}^{K} {\parallel b_k^{prior} - b_k \parallel}_{\Gamma_{k}}^{2} 
    + \sum_{(t,k) \in G}^{J_{\Delta}} {\parallel h(m_t, b_k) - g_{tk} \parallel}_{\Omega_{j}}^{2} \\
    + \sum_{(t,s) \in Q}^{L_{\Delta}} {\parallel d(m_t, r_s) - q_{ts} \parallel}_{\Xi_l}^{2}  
    + \sum_{v \in V}^{W} {\parallel r_v^{prior} - r_v \parallel}_{\Lambda_l}^{2}
\end{multline}

We apply iSAM2 \cite{isam2} to efficiently solve Eq. \ref{eq:map} online.

\section{EXPERIMENTAL SETUP}

\subsection{SAM AUV}
The vehicle used in the experiments is the SAM AUV (bottom of Fig. \ref{fig:farm_real}), introduced in \cite{Bhat_etal:2020b}. 
Its navigation sensors are a doppler velocity log (DVL), compass, IMU and control feedback, its middleware is ROS \cite{quigley2009ros} and its payload SSS is a hull-mounted DeepVision DE3468D with a sampling frequency of 10Hz. 




\begin{figure*}[t]
    \vspace{0.075in}
    \makebox[\textwidth]{\hspace{.1in}(V)\hspace{.55in}(I)\hspace{1.25in}(III)\hspace{.55in}(IV)\hspace{1.4in}(II)}\\
    \centering
    \includegraphics[width=0.75\linewidth]{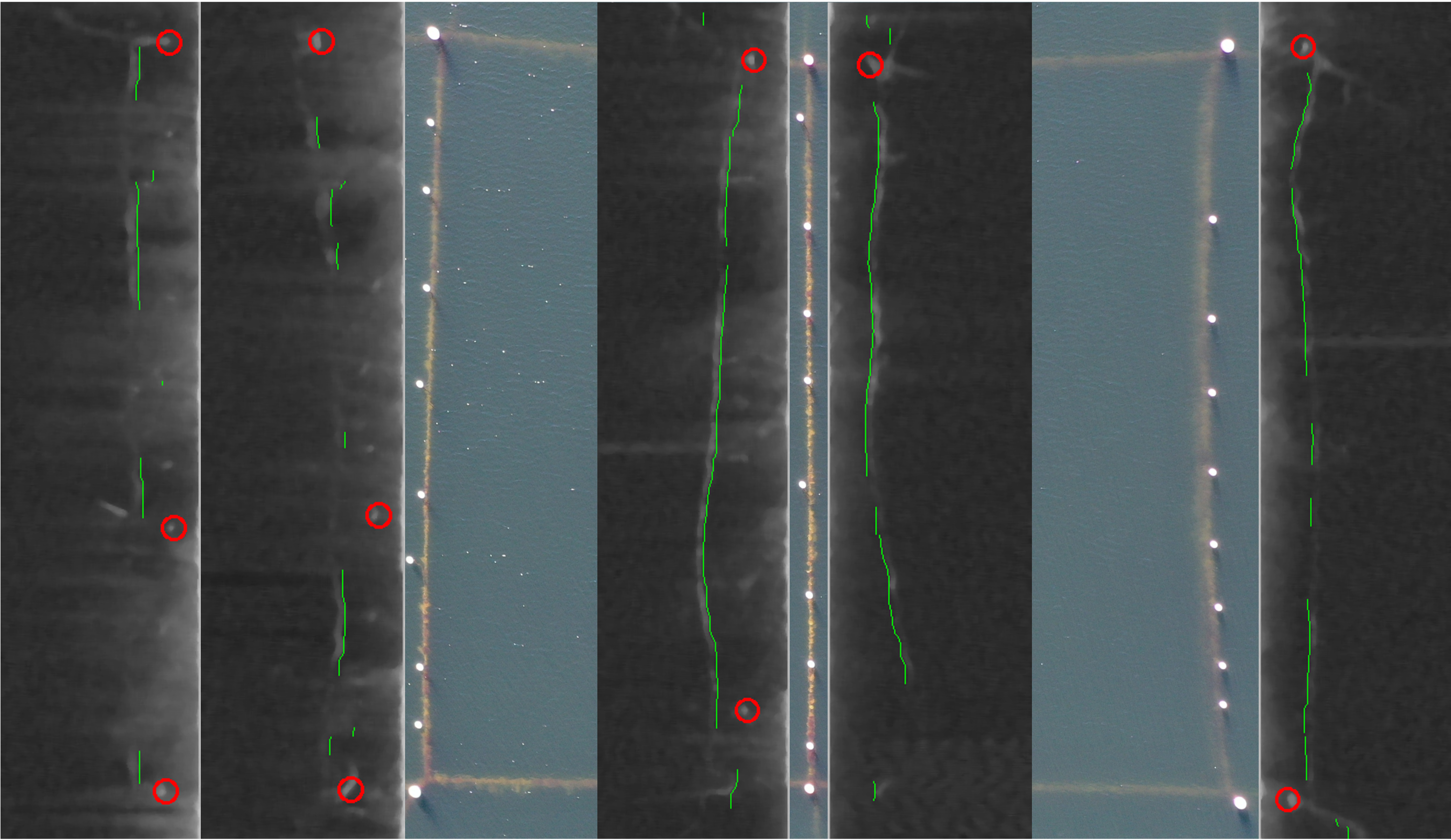}
    \caption{SSS detector results referenced to the aerial image of the algae farm. The numerals correspond to the pass number of the AUV survey, in Fig. \ref{fig:farm_real}. Red markers show buoy and green, individual rope detections. Note the missed buoy on IV.}
    \label{fig:detector_overlay}
\end{figure*}

\subsection{AUV survey of a kelp farm}

A survey from the surface was carried out on the algae farm at the top of Fig. \ref{fig:farm_real} at the Kristineberg marine field station off the west coast of Sweden. The farm consists of three lines of algae of approximately 26 meters separated by two corridors 13 meters wide, mooring at a depth of approximately 1.5 meters.
The survey consisted of five swaths parallel to the lines at approximately 2 meters distance, with two swaths in between the lines and the other three on the external sides of the outer ropes (see Fig. \ref{fig:detector_overlay}). It covered a distance of approximately 300 meters in $T = 430$ seconds. The challenging water conditions of the day can be seen in the quickly degrading DR estimate of the vehicle, in red in Fig. \ref{fig:final_trajectories}.
Due to the on-board GPS malfunctioning during the survey, no initial GPS fix could be recorded, and therefore the coordinates of the auxiliary ship that followed the AUV during the mission were used to initialize the framework.


\section{RESULTS}
\label{sec:results}



\subsection{SLAM evaluation}
We assess the performance of the presented framework in the tasks of online vehicle localization and environment mapping on the data collected at the algae farm. The survey was replayed in real time in a HIL setup in a Jetson Orin. We compare the back-end methods in \cite{zhang2023fully} (baseline 1) and \cite{fallon2011efficient} (baseline 2) against our own to evaluate in detail the contributions of the different parts of the system. Baseline 1 failed to produce usable feature matches without an entirely tailor-made front-end and was therefore limited to use the buoy detections as constraints for the back-end.

\begin{figure*}[t]
\vspace{0.075in}
\centering
\begin{subfigure}[b]{.32\linewidth}
\centering
		\includegraphics[width=\linewidth]{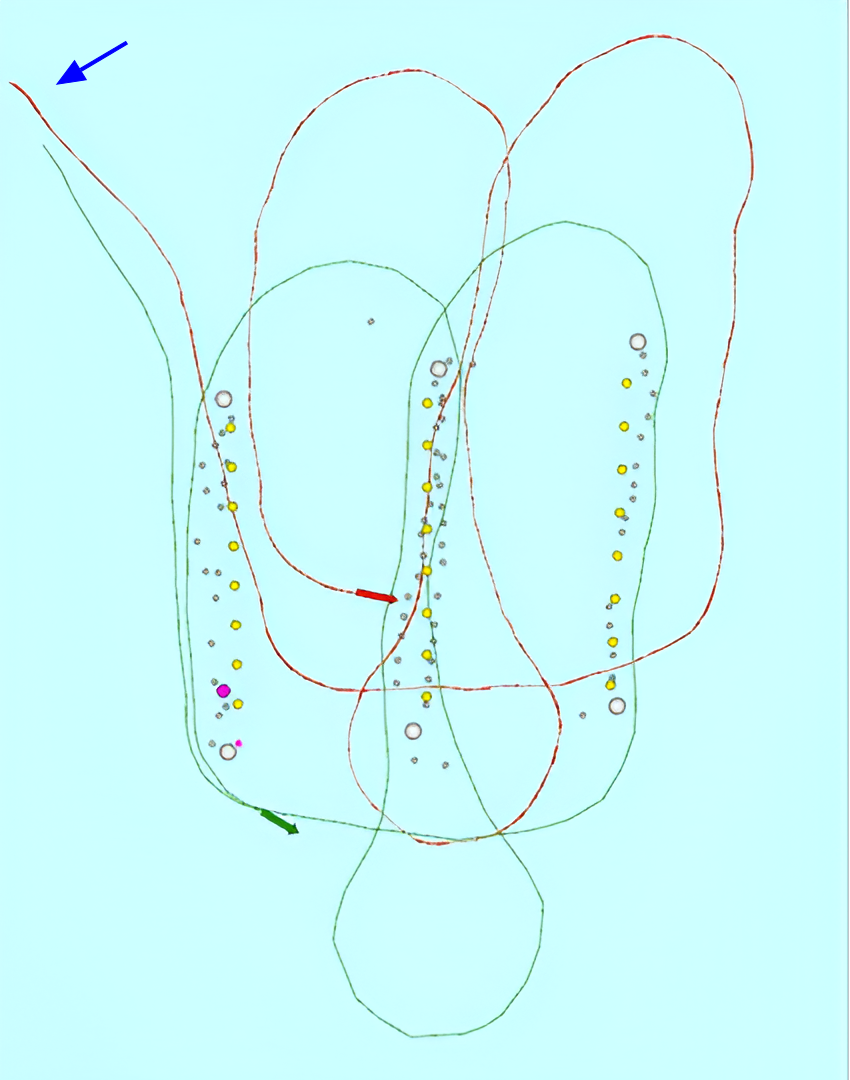}
\caption{Baseline 1 \cite{zhang2023fully}}
\label{subfig:method 3 final}
\end{subfigure}
\begin{subfigure}[b]{.32\linewidth}
\centering
	\includegraphics[width=\linewidth]{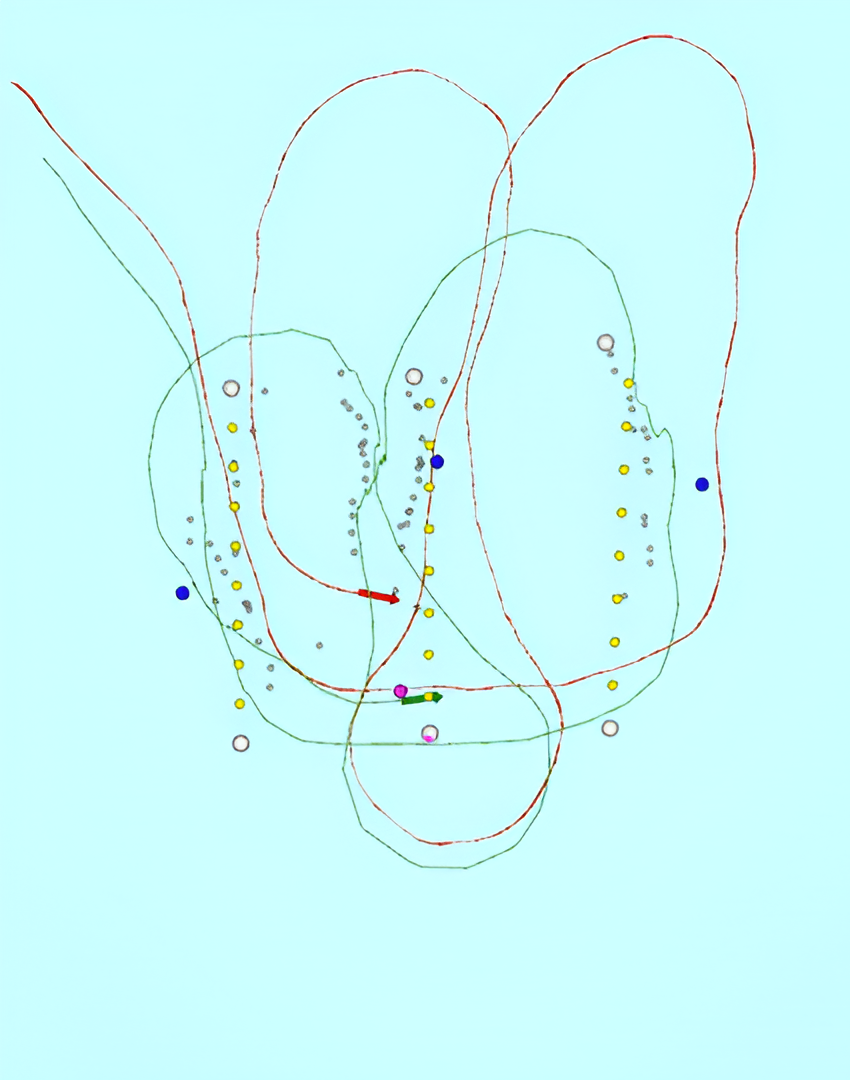}
\caption{Baseline 2  \cite{fallon2011efficient}}
\label{subfig:method 1 final}
\end{subfigure}
\begin{subfigure}[b]{.32\linewidth}
\centering
        \includegraphics[width=\linewidth]{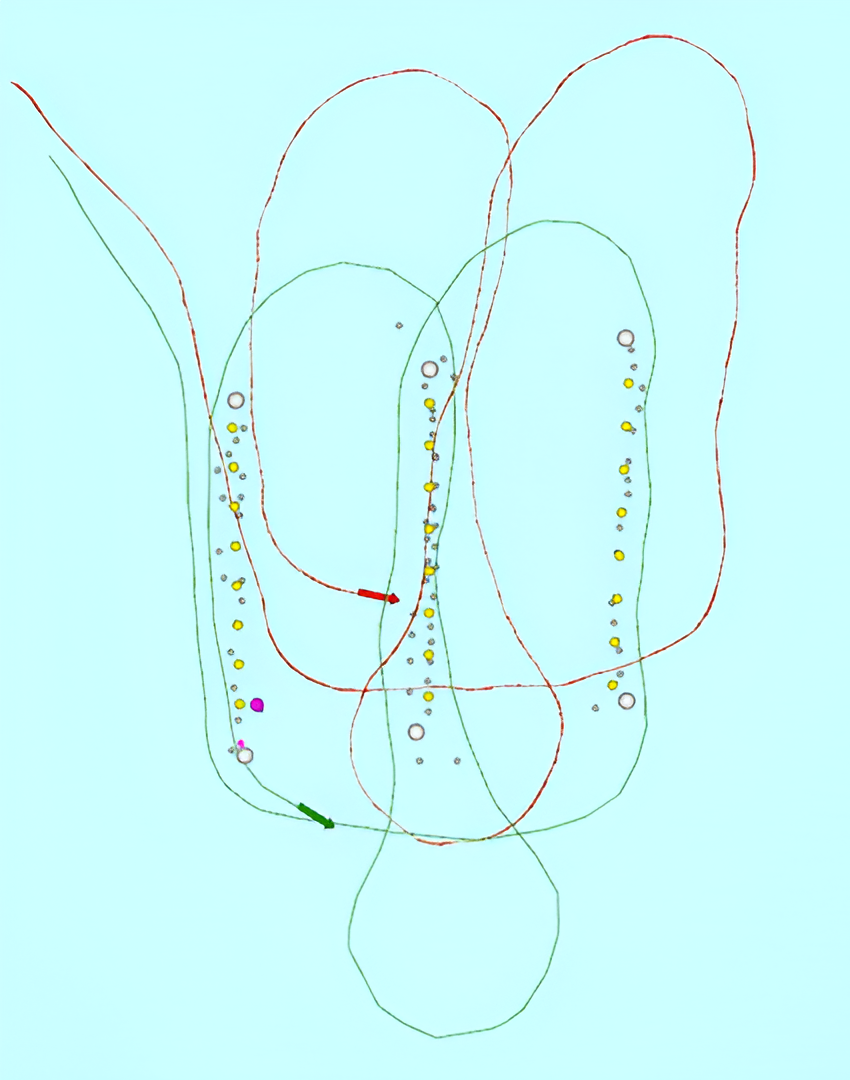}
\caption{Proposed method}
\label{subfig:method 2 final}
\end{subfigure}

\caption{RVIZ depiction of the final AUV optimized (green) and DR (red) estimates, mapped buoys (large white circles), intermediate buoys (yellow circles) and mapped rope detections (gray circles) of the methods compared. The blue arrow indicates the start of the survey. Full visualization available at \url{https://github.com/julRusVal/sss_farm_slam}. }
\label{fig:final_trajectories}
\end{figure*} 


\subsubsection{Mapping evaluation}
Given the lack of ground truth measurements of the actual location of the ropes, we can only directly evaluate the outcome of the SLAM methods comparing the buoys final estimates $B^*_t$ from Eq. \ref{eq:map} against their priors $B_0$. However, we further measure how well the final rope detections $R^*_t$ fit lines for each method. These give us a reasonable estimate of both the mapping accuracy and the final AUV trajectory estimate to gauge performances. The root mean squared errors (RMSE) for both sets of variables can be seen in the first two columns of Table \ref{tab:rope_error}.

\begin{table}[h]
\centering
\begin{tabular}{c|ccc|}
\cline{2-4}
\multicolumn{1}{l|}{}        & \multicolumn{3}{c|}{RMSE (m)}                                                                  \\ \hline
\multicolumn{1}{|c|}{Method} & \multicolumn{1}{l|}{Rope} & \multicolumn{1}{l|}{Buoy} & \multicolumn{1}{l|}{oRPE} \\ \hline
\multicolumn{1}{|c|}{Baseline 1 (\cite{zhang2023fully})}      & 1.23                      & 1.06                      & 2.68                                   \\ \hline
\multicolumn{1}{|c|}{Baseline 2 (\cite{fallon2011efficient})}      & 2.55                      & 0.95                      & 5.53                                   \\ \hline
\multicolumn{1}{|c|}{Proposed}      & 1.00                      & 1.14                      & 2.04                                   \\ \hline
\end{tabular}
    \caption{Final map and trajectory RMS errors by method.}
    \label{tab:rope_error}
\end{table}

\subsubsection{Trajectory estimate evaluation}
Fig. \ref{fig:final_trajectories} shows the final estimates for each method $M^{*, final}$ (green) together with the DR estimate (red) over the representation of the farm in RVIZ \cite{quigley2009ros}. The only qualitative information available on the real AUV trajectory is that it never crossed the algae ropes. Based on that it is easy to see in Fig. \ref{fig:final_trajectories} that baseline 2 has not yielded an accurate final estimate.
However, both baseline 1 and our method produce realistic full trajectories. 
\begin{figure}[h]
    \centering
    \includegraphics[width=0.90\linewidth]{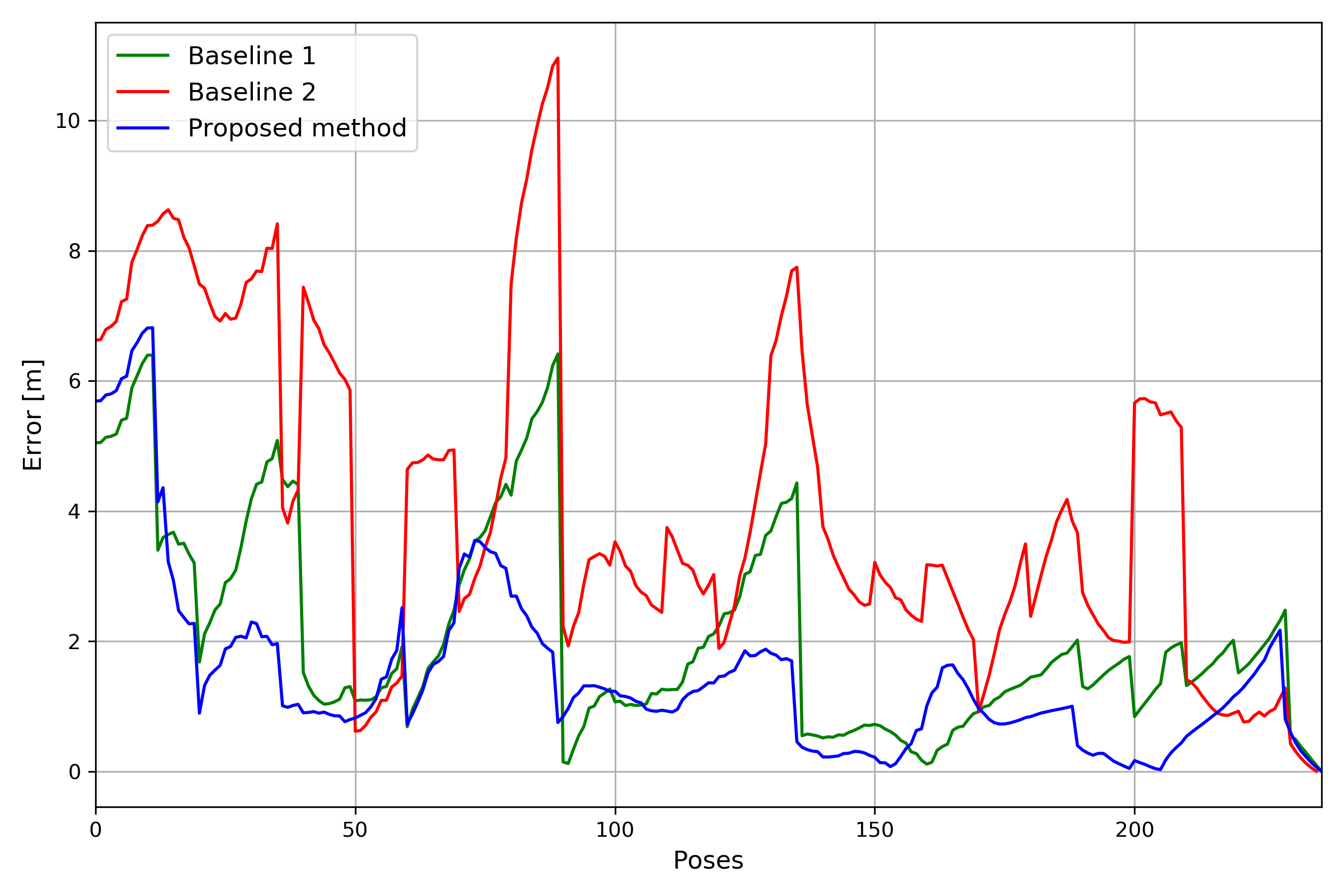}
    \caption{Online relative pose error (oRPE) of the trajectory estimates per method.}
    \label{fig:online_error_individual}
\end{figure}

Not having the ground truth trajectory of the vehicle, we must therefore compare them in relative terms. To do so, we assume that both methods yield a close-to-optimal final trajectory estimate at the end of the mission, $M^{*, final}$. We then analyze the smoothness of the error between the incremental solution at each time step, named $m^{*, online}_\Delta$ and the corresponding value from the final solution $m^{*, final}_\Delta$.
The resulting "online relative pose error" can be formulated as $oRPE_\Delta = m^{*, online}_\Delta - m^{*, final}_\Delta$ and is plotted in Fig. \ref{fig:online_error_individual} for every method. Table \ref{tab:rope_error} summarizes the RMSE results.


Fig. \ref{fig:online_error_individual} shows how the over-reliance of baseline 1 on the buoy detections yields an online estimate that diverges greatly between buoy measurements. However, the use of rope detections in the proposed method laterally constrained the online estimate through the mission, producing a smoother evolution of the error over time and an overall smaller RMS of the oRPE, in Table \ref{tab:rope_error}. 

\subsection{SLAM time performance evaluation}
All the experiments have been executed in the payload computer of the SAM AUV, a Jetson Orin AGX 64. The rope and buoy detectors used in the front-end were capable of running at $\sim11$ Hz. The statistics on the back-end optimization performance, shown in Table \ref{tab:time_performance_stats}, indicate that the proposed method incurs a small time penalty compared to baseline 1 from the integration of rope measurements. Despite this, the method remains suitable for real-time operation on and AUV when compared to the full survey duration $T$.  



\begin{table}[h]
    \centering
    \begin{tabular}{|c|c|c|c|}
        \hline
        Method & Max. update time (s) & Total time (s)  & Factor count\\ \hline
        Baseline 1 & 0.30   & 15.0 & 254   \\ \hline
        Baseline 2 & 0.26   & 15.3 & 322   \\ \hline
        Proposed   & 0.29   & 17.0 & 381   \\ \hline
    \end{tabular}
    \caption{Comparison of iSAM2 update times and corresponding factor graph sizes across different methods.}
    \label{tab:time_performance_stats}
\end{table}

\section{CONCLUSIONS}
We have presented a novel SLAM method for a low-cost AUV equipped with a SSS in an algae farm. Key to our SLAM approach is the modeling of the algae line detections from SSS pings as individual landmarks with their own instances of rope priors, which match the priors of the originating ropes. We have shown experimentally that this treatment of the ropes yields sparsely-connected graphs whose solutions succeed in maintaining a smoother AUV pose estimate along the ropes, resulting in more accurate online AUV trajectory estimate than the baseline methods compared.



Overall, the presented HIL results, although limited to one dataset, show the amenability of our framework to on-board, autonomous navigation and surveying of algae farms, a first step towards fully automated marine farming.

\section*{ACKNOWLEDGMENT}
 This work was financially supported by the Swedish Foundation for Strategic Research through the Swedish Maritime Robotics Center (IRC15-0046) and the Vinnova project: Underwater Drones for Seaweed Farm Monitoring and Sampling (dnr. 2020-04551). We thank Fredrik Gröndal for providing access to the algae farm for this research.

\bibliographystyle{IEEEtran}
\bibliography{IEEEabrv,root}

\end{document}